\pdfoutput=1
\documentclass[11pt]{article}

\usepackage[margin=1in]{geometry}
\usepackage{amsmath,amssymb}
\usepackage{booktabs}
\usepackage{enumitem}
\usepackage{hyperref}
\usepackage{xcolor}
\usepackage{microtype}
\usepackage{parskip}
\usepackage{graphicx}

\title{\textbf{A Universal Context-Reuse Layer for Cross-Model KV Sharing}}

\author{Yi Li, Dongming Jiang, Yi Zhao, and Bingzhe Li\\
University of Texas at Dallas}

\date{}

\begin{document}

\maketitle

\begin{abstract}
Modern large language model (LLM) serving systems increasingly operate over repeated or shared context, yet each model typically performs its own prefill computation even when another model has already processed the same input. Existing KV-cache reuse mechanisms substantially reduce redundant computation within a single model, but generally assume that the producer and consumer of a cache are identical. We study \emph{cross-model KV sharing}, which translates the KV state produced by a source model into a representation that can be consumed by a different target model, including models that differ in scale, architecture, attention configuration, tokenizer, and model family. We evaluate the approach in both within-family and cross-family settings. For Qwen2.5-7B $\rightarrow$ Qwen2.5-1.5B, translated KV states improve LongBench2 accuracy from 27.59\% to 34.48\%, a gain of 6.89 percentage points over the native 1.5B baseline, while reducing handoff cost relative to native target prefill. For the cross-family Qwen2.5-1.5B $\rightarrow$ Gemma-2-2B setting, KV handoff reduces target-side prefill cost by up to 67.05\% at 4K context length while maintaining decoding perplexity close to native-model baselines. In a more heterogeneous Llama3.1-70B $\rightarrow$ Qwen2.5-7B setting, cross-family handoff achieves 44.0\% accuracy compared with 45.7\% for native Qwen2.5-7B inference, while reducing measured latency from 899ms to 138ms. These results provide initial evidence that KV states can serve as transferable computational representations rather than strictly model-local caches, and motivate \emph{context mobility} as a systems abstraction for reducing redundant prefill across heterogeneous LLM and multi-agent inference workflows.

\end{abstract}

\section{Introduction}

Large language model (LLM)~\cite{vaswani2017attention, touvron2023llama, brown2020language} inference is increasingly becoming a systems problem involving not one model, but a collection of models that cooperate within the same application. Modern AI services increasingly compose multiple LLMs within a single application. For example, model-routing and cascading systems dynamically select among models with different capability--cost profiles, often sending easier requests to smaller models and escalating more difficult requests to stronger models~\cite{ong2025routellm, mohammadshahi2024routoo}. Other systems combine specialized models for different tasks or coordinate multiple LLM-based agents that communicate and operate over shared task state, conversation history, or retrieved information~\cite{guo2024large, wang2024survey}. Such heterogeneity is attractive because available models exhibit different tradeoffs in accuracy, latency, inference cost, task specialization, and deployment constraints.

However, current inference infrastructure does not efficiently exploit the fact that these heterogeneous models often operate over the same underlying context. When a new model is invoked, it normally processes the input tokens independently, even if another model has already processed exactly the same conversation history, document collection, or user prompt. As a result, the same semantic content may be repeatedly transformed into model-specific internal representations.

This redundancy is particularly important during the \emph{prefill} stage of LLM inference. During prefill, the model processes all input tokens through its transformer layers and constructs the key-value (KV) states required by the attention mechanism. For a long prompt, this process can involve substantial GPU computation, memory traffic, and latency before the first output token is generated. Once the KV states have been computed, subsequent decoding can reuse them efficiently.

Existing KV caching and prefix-caching systems already exploit this observation within a single model. If the same model encounters a prefix that has already been processed, the system can retrieve its existing KV cache rather than recomputing the prefix. This optimization has become an important component of modern LLM serving infrastructure.

The limitation is that KV states are conventionally treated as model-specific objects. A KV cache generated by one model is generally unusable by another model because the models may have different hidden representations, attention configurations, numbers of layers, tokenizers, and architectural parameters.

This paper investigates a broader form of reuse: \textbf{cross-model KV sharing}. The central idea is to translate the KV representation generated by one model into a representation that can be consumed by another model. The source and target models may be different-sized variants within the same family or, more generally, models belonging to entirely different architectural families.

The central systems principle is:

\begin{quote}
\textbf{If multiple models operate over the same context, the computational work used to understand that context should not necessarily be repeated independently by every model.}
\end{quote}

Cross-model KV sharing therefore seeks to make contextual computation partially portable across model boundaries. Rather than treating each model invocation as an isolated computation beginning from raw tokens, the system attempts to reuse useful internal state produced by earlier models.

This capability is especially relevant to emerging heterogeneous and agentic AI systems, where model transitions are increasingly common and shared context can be long-lived. A short video demonstration illustrating cross-model KV sharing in practice is available
\footnote{Demo video: \url{https://youtu.be/0TiOdUK0qB8}}.

\section{Background and Related Work}

\subsection{KV Caching and Context Reuse}

Transformer-based LLMs rely on self-attention to condition each token on preceding tokens in the sequence. At each transformer layer, the model computes query, key, and value representations. During autoregressive generation, the key and value representations associated with previously processed tokens can be retained in a key-value (KV) cache and reused during subsequent decoding steps.

Consider an input context

\[
C = (x_1,x_2,\ldots,x_L),
\]

where \(L\) denotes the number of input tokens. During prefill, an LLM \(M\) processes the complete sequence and generates a collection of layer-wise KV states,

\[
KV_M(C)
=
\left\{
(K^{(1)},V^{(1)}),
\ldots,
(K^{(N)},V^{(N)})
\right\},
\]

where \(N\) is the number of transformer layers. Once these states have been constructed, subsequent decoding can reuse them rather than recomputing the representations of all preceding tokens. KV caching therefore converts the computation performed during prefill into reusable runtime state.

For long-context workloads, constructing this state can represent a substantial computational cost. Processing thousands or tens of thousands of input tokens requires significant matrix computation, attention processing, memory movement, and, in distributed deployments, inter-device communication before decoding begins. Consequently, KV-cache management has become an important component of modern LLM-serving systems.

\subsection{Related Work}
A broad class of serving optimizations exploits this opportunity. Prefix caching reuses KV states when multiple requests share a common prefix; paged KV management improves memory utilization and allocation flexibility; cache offloading and distributed KV storage extend effective cache capacity beyond local accelerator memory; and cache-aware scheduling attempts to colocate requests with reusable state. These techniques differ in how KV states are stored, placed, transferred, and scheduled, but they primarily optimize reuse of states that remain semantically and structurally associated with the model that originally generated them.

Prior work on KV-cache reuse spans several complementary directions~\cite{yao2025cacheblend, ye2026kvcomm, liu2026droidspeak, hu2026swiftcache, wang2025kvcache, fang2026flashagents}. Conventional prefix caching and modular prompt-reuse systems avoid redundant prefill when identical or structurally compatible context reappears within the same model. Representative systems include prompt caching and RAG-oriented cache managers, which treat reusable prompt segments or retrieved documents as cacheable computational units.

CacheBlend~\cite{yao2025cacheblend} extends KV reuse beyond exact-prefix matching by allowing independently cached text chunks to be combined under a new context. Because independently cached chunks may not encode the cross-chunk attention dependencies required by the newly composed prompt, CacheBlend selectively recomputes a subset of token states while reusing the remaining KV cache. This line of work demonstrates that previously computed KV states can remain useful even when the surrounding context changes, provided that the resulting contextual mismatch is appropriately corrected.

KVCOMM~\cite{ye2026kvcomm} further studies cross-context KV reuse in LLM-based multi-agent systems, where different agents may repeatedly process overlapping content under different prefix contexts. It introduces a training-free mechanism that estimates and corrects context-induced KV-cache offsets using an online set of anchor examples, allowing shared content to be reused despite changes in preceding context. Both CacheBlend and KVCOMM therefore broaden KV reuse beyond exact-prefix matching, while primarily addressing \emph{context mismatch} within the same underlying model.

Another line of work focuses on KV-cache placement and transport rather than representation conversion. SwiftCache~\cite{hu2026swiftcache} exploits underutilized GPU memory and high-bandwidth interconnects across heterogeneous model-serving workers to improve KV-cache storage and retrieval. Such systems make KV states increasingly mobile across hardware resources, but the cached representation itself remains associated with the model that originally generated it; moving a cache does not make it directly consumable by another model with a different internal representation.

More closely related to cross-model reuse, DroidSpeak~\cite{liu2026droidspeak} studies KV-cache sharing across fine-tuned model variants that use the same underlying architecture. It identifies layers whose KV states remain sufficiently compatible across model variants and selectively recomputes layers whose representations differ more substantially. DroidSpeak therefore demonstrates that KV reuse can extend beyond identical model instances, although its design relies on strong architectural correspondence between the source and target models.

Recent work further investigates explicit KV translation between models of different sizes. NVIDIA~\cite{heo2026cross} study KV-cache transfer within LLM families and show that useful linear relationships can exist between source- and target-model KV representations. Their approach constructs closed-form linear mappings for predicting target KV states from source-model caches, enabling prefill reuse during model cascading or switching. The method primarily targets structurally related source--target pairs within the same model family.

Heterogeneous KV translation extends this idea across architectural families. Mixture-of-Translators (MoT)~\cite{lee2026mixture} formulates cross-model KV reuse as a learned representation-translation problem and employs multiple translator modules to map cached states across heterogeneous LLMs. This work provides evidence that KV representations can be translated even when the participating models do not share the same architecture, moving beyond direct cache reuse and within-family mappings.

Collectively, these studies expand KV reuse along several dimensions. CacheBlend~\cite{yao2025cacheblend} and KVCOMM~\cite{ye2026kvcomm} address changes in the context surrounding cached representations; SwiftCache~\cite{hu2026swiftcache} improves the physical storage and movement of KV states; DroidSpeak~\cite{liu2026droidspeak} enables reuse across architecturally aligned model variants; Heo et al.~\cite{heo2026cross} translate KV states across different model scales within a family; and MoT~\cite{lee2026mixture} investigates translation across heterogeneous model architectures. Our work builds on this progression by studying both within-family and cross-family KV sharing as a mechanism for preserving previously computed contextual state across heterogeneous inference workflows.

\section{The Redundant Context Computation Problem}

Consider an AI workflow in which several models operate over a common context \(C\). For example:

\[
C
\rightarrow
M_1
\rightarrow
M_2
\rightarrow
M_3.
\]

The context may include a system prompt, user request, conversation history, retrieved documents, tool outputs, or intermediate agent state.

Under conventional inference, each model independently performs its own prefill computation:

\[
KV_{M_1}(C)
=
\mathrm{Prefill}(M_1,C),
\]

\[
KV_{M_2}(C)
=
\mathrm{Prefill}(M_2,C),
\]

\[
KV_{M_3}(C)
=
\mathrm{Prefill}(M_3,C).
\]

The aggregate prefill cost is approximately

\[
C_{\mathrm{native}}
=
\sum_{i=1}^{N}
C_{\mathrm{prefill}}(M_i,C).
\]

From the perspective of each individual model, this computation is necessary because the model requires its own internal representation. From the perspective of the overall application, however, the system is repeatedly processing substantially the same information.

The inefficiency becomes more severe as either the context length or the number of participating models increases. A short 100-token request may make redundant prefill relatively unimportant. In contrast, a 32K-token conversation history repeatedly processed by several large models can represent significant duplicated computation.

The same problem appears across multiple emerging workloads. In \textbf{model escalation}, a small model may initially handle a request before escalating it to a larger model, which must then recompute the entire context. In \textbf{multi-agent systems}, several agents using different models may consume the same user history and retrieved evidence. In \textbf{model routing}, requests may move dynamically among models based on task difficulty, cost, latency, or specialization. Similarly, in \textbf{verification pipelines}, one model may generate an answer while another independently processes the same input to verify or critique it. Across all these cases, the logical information is shared, but the underlying computational representations are not. Cross-model KV sharing seeks to bridge this gap.

\begin{figure*}[t]
  \centering
\includegraphics[width=0.7\textwidth]{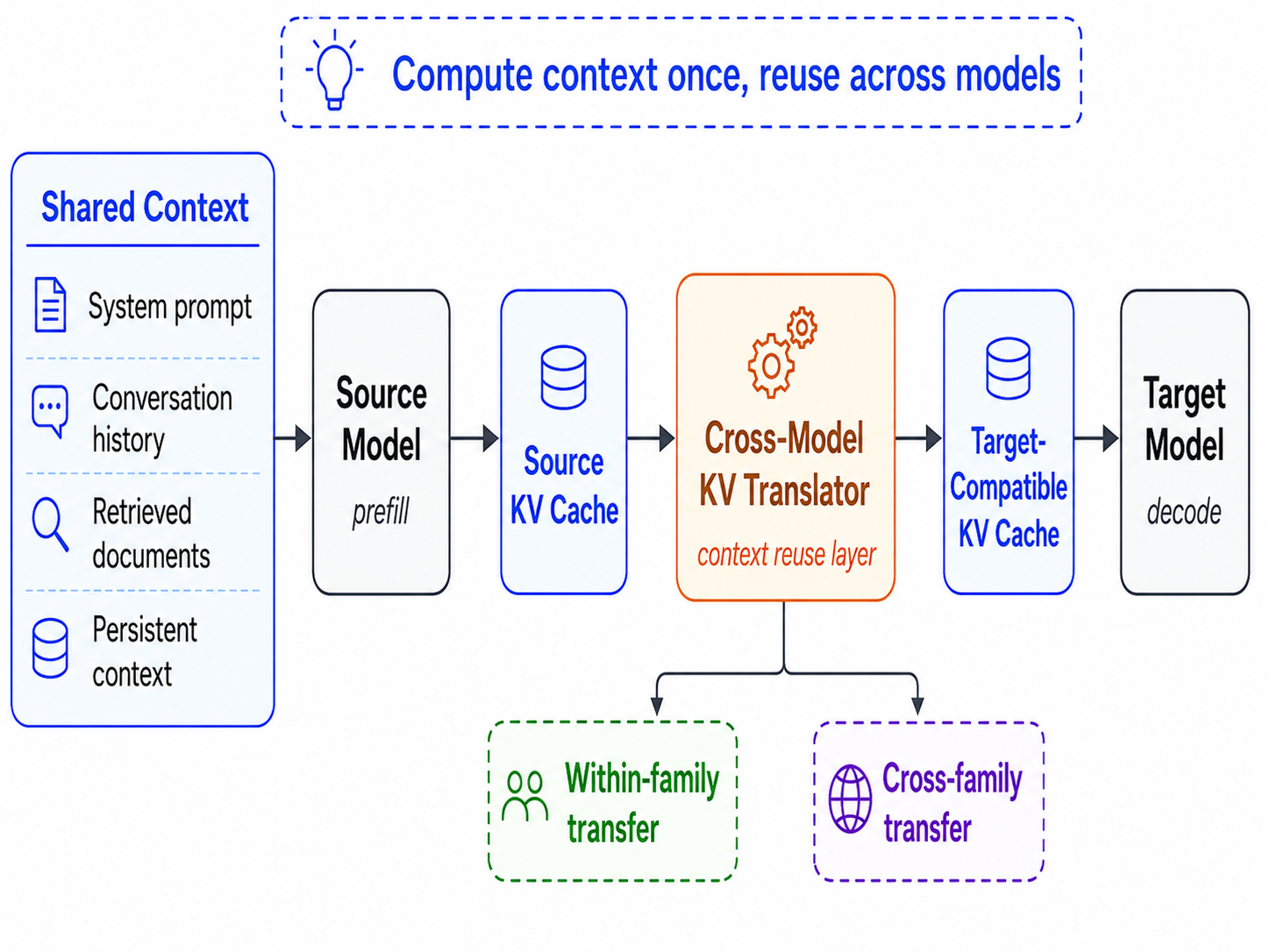}
  \caption{Overall view of cross-model KV sharing.}
  \label{fig:overall}
\end{figure*}
\section{Cross-Model KV Sharing}

Cross-model KV sharing enables one language model to reuse contextual computation that was previously performed by another model. In conventional inference, each model independently processes the same input context and constructs its own KV cache, even when another model has already processed that context. For a shared context \(C\), this means that different models separately compute their own representations:$
KV_A(C)$ and $KV_B(C)$.
This repeated prefill computation becomes increasingly inefficient in systems that involve multiple models operating over shared prompts, conversation histories, retrieved documents, or other persistent context.

The key idea is to treat the KV cache not only as a model-local runtime artifact, but as a potentially reusable computational representation. Rather than requiring a target model to independently reconstruct the entire context from the original tokens, cross-model KV sharing introduces a translation mechanism that converts the source model's KV representation into a target-compatible representation:

\[
\widehat{KV}_B(C)
=
T_{A\rightarrow B}\bigl(KV_A(C)\bigr),
\]

where \(T_{A\rightarrow B}\) denotes the transformation from the source model's representation space to that of the target model. The resulting translated state can then be used to support subsequent inference in the target model.

This translation is more than a tensor-format conversion or dimensional reshaping. The source KV cache reflects how the source model encodes contextual relationships across its attention layers, while the target model may use a different learned representation space, attention structure, layer organization, hidden dimension, or tokenizer. A successful translation must therefore preserve the contextual information that is useful for subsequent inference while adapting it to the computational structure expected by the target model.

Importantly, the translated representation does not necessarily need to reproduce the target model's native KV cache exactly. In general,

\[
\widehat{KV}_B(C) \neq KV_B(C)
\]

may still be acceptable. The practical objective is to construct a representation that provides sufficient contextual information for the target model to continue inference with acceptable output quality while avoiding a meaningful portion of target-side prefill computation. This creates a design space in which different translation mechanisms can balance reconstruction fidelity, translation overhead, runtime efficiency, generalization, and compatibility across heterogeneous models.

Cross-model KV sharing therefore generalizes conventional KV-cache reuse beyond the assumption that the producer and consumer of a cache must be the same model. The source and target models may differ in model size, configuration, architecture, tokenizer, or model family. As these differences increase, the translation problem becomes more challenging because the underlying representations are less directly aligned. Nevertheless, the fundamental objective remains the same: preserve as much as possible of the computational value already invested in processing the context and carry that value across model boundaries.

From a systems perspective, cross-model sharing changes the execution model of heterogeneous AI applications. Instead of requiring every newly invoked model to restart from raw tokens, the runtime can first determine whether relevant contextual computation already exists and whether an appropriate translation path is available. When reuse is beneficial, the target model can begin from a translated KV state rather than repeating the full prefill process. When the translated representation does not meet the required quality or efficiency criteria, the system can fall back to conventional native prefill.

The efficiency benefit can be expressed at a high level by comparing the cost of translation with the cost of native prefill. Cross-model reuse is advantageous when

\[
C_{\mathrm{trans}} + C_{\mathrm{overhead}}
<
C_{\mathrm{prefill}},
\]

provided that the translated state also satisfies the application's quality requirements. The potential benefit therefore increases as prompts become longer, target models become more computationally expensive, and applications involve more frequent transitions among models.

More broadly, cross-model KV sharing introduces a different way to view contextual computation in LLM systems. KV states need not be treated solely as temporary, model-specific artifacts associated with individual inference calls. Instead, they can be viewed as reusable computational assets that may be stored, transferred, translated, and consumed across different models. This perspective provides the foundation for a broader notion of \emph{context mobility}, in which the computational value of previously processed information can persist as an AI workload moves across heterogeneous models, agents, and execution environments.

\section{Understanding the Translation Layer}

The translation layer is the central component that enables cross-model reuse.

At a high level, the system observes source KV states

\[
KV_A(C)
\]

and learns or constructs a function that produces a representation suitable for a target model.

The general objective can be written as

\[
\widehat{KV}_B(C)
=
T_{\theta}
\left(
KV_A(C)
\right),
\]

where \(\theta\) represents the parameters of the translation mechanism.

One possible learning objective is to minimize the difference between the translated state and a target state generated by native prefill:

\[
\mathcal{L}_{KV}
=
D
\left(
T_{\theta}(KV_A(C)),
KV_B(C)
\right),
\]

where \(D(\cdot,\cdot)\) is an appropriate representation-distance function.

However, minimizing raw KV distance may not be the only or even the most important objective. Two KV states can differ numerically while still leading to similar model behavior. A more task-oriented objective may therefore optimize downstream outputs, attention behavior, logits, or task accuracy.

This perspective is important because cross-model sharing does not aim to reproduce the exact internal computation that the target model would have generated; rather, it seeks to construct a compatible state that enables the target model to continue inference effectively. Such translation can operate across individual layers, groups of layers, individual KV heads, shared intermediate representations, compressed KV representations, or low-dimensional latent adapters. The translation mechanism can therefore be designed to balance translation fidelity, runtime overhead, memory footprint, and the ease of supporting additional models.

\begin{table}[t]
    \centering
    \caption{LongBench2 accuracy for within-family Qwen2.5 KV sharing.}
    \label{tab:qwen-within-family-acc}
    \begin{tabular}{lc}
        \toprule
        \textbf{Inference mode} & \textbf{Accuracy} \\
        \midrule
        Qwen2.5-7B self-inference & 45.69\% \\
        Qwen2.5-1.5B self-inference & 27.59\% \\
        7B $\rightarrow$ 1.5B KV handoff & \textbf{34.48\%} \\
        \bottomrule
    \end{tabular}
\end{table}
\section{Experimental Evaluation}
\label{sec:evaluation}

We evaluate cross-model KV sharing under progressively more heterogeneous source--target pairs. Our experiments ask two questions: (1) whether KV states can be transferred between differently sized models from the same family, and (2) whether useful state can still be transferred when the models differ in architecture, tokenizer, and scale.

\subsection{Experimental Setup}
\label{sec:experimental-setup}

\paragraph{Model configurations.}
We study one within-family and two cross-family transfer directions. The within-family experiment uses Qwen2.5-7B-Instruct as the source and Qwen2.5-1.5B-Instruct as the target. The cross-family experiments use Qwen2.5-1.5B-Instruct $\rightarrow$ Gemma-2-2B-IT, which isolates family-level heterogeneity at a comparable parameter scale, and Llama3.1-70B $\rightarrow$ Qwen2.5-7B, which combines family-level heterogeneity with a large source--target scale gap.

\paragraph{Handoff protocol.}
For each configuration, the source model prefills the prompt and produces its native KV cache. A learned transport module maps this cache into the target model's KV representation space, after which the frozen target model decodes directly from the translated state. The target therefore does not independently prefill the full prompt. We compare this handoff path with source self-inference and target self-inference, in which the corresponding model performs both prefill and decoding. All handoff latency results measure the incremental cost of reusing an already available source cache; they do not include source prefill and should not be interpreted as the cost of generating a cache solely for transfer.

\paragraph{Datasets and metrics.}
For the within-family experiment, the transport module is trained on prompts from NoLiMa and evaluated on 116 scored LongBench2 samples. We report multiple-choice accuracy over the A/B/C/D options using log-probability scoring, both overall and by context-length bucket. Latency is measured over three samples in each bucket, and we separately instrument translation, peer-to-peer copy, and cache assembly. For Qwen2.5-1.5B $\rightarrow$ Gemma-2-2B, we use text sampled from FineWeb-Edu and report native prefill and handoff latency for prompts from 128 to 4K tokens. We also report perplexity over decoding horizons $H\in\{1,16,32,128,512\}$, where lower values are better. For Llama3.1-70B $\rightarrow$ Qwen2.5-7B, we report multiple-choice accuracy based on log-probability scoring and latency for native and handoff execution.

\subsection{Within-Family Results}
\label{sec:within-family-results}

\paragraph{Accuracy.}
Table~\ref{tab:qwen-within-family-acc} compares native inference with Qwen2.5-7B and Qwen2.5-1.5B against the translated-cache handoff. Decoding with the translated 7B cache improves the frozen 1.5B target from 27.59\% to 34.48\%, an absolute gain of \textbf{6.89 percentage points}. Because the target parameters remain unchanged, this gain is attributable to information carried by the translated KV state. The handoff remains below the 45.69\% accuracy of native 7B inference, as expected from the lower capacity of the target; nevertheless, it recovers 38.1\% of the accuracy gap between native 1.5B and 7B inference.

\paragraph{Effect of context length.}
Table~\ref{tab:qwen-length} partitions handoff accuracy by prompt length. Accuracy is 40.74\% for 8K--16K contexts and 32.58\% for 16K--32K contexts. Thus, the translated state remains informative at up to 32K tokens, but accuracy decreases by 8.16 percentage points in the longer bucket. This gap suggests that preserving representation fidelity becomes more difficult as the transported context grows and motivates transport mechanisms designed explicitly for long-context alignment.

\begin{table}[t]
    \centering
    \caption{Within-family handoff accuracy by context length.}
    \label{tab:qwen-length}
    \begin{tabular}{lcc}
        \toprule
        \textbf{Context length} & \textbf{Samples} & \textbf{Accuracy} \\
        \midrule
        8K--16K  & 27 & 40.74\% \\
        16K--32K & 89 & 32.58\% \\
        \bottomrule
    \end{tabular}
\end{table}

\paragraph{Handoff efficiency.}
As shown in Table~\ref{tab:qwen-timing}, translating an existing source cache is substantially cheaper than native prefill. Handoff takes 34.5~ms for 8K--16K contexts, compared with 158.7~ms for 1.5B prefill, yielding a 4.6$\times$ reduction in incremental latency. For 16K--32K contexts, handoff takes 53.8~ms versus 288.3~ms for target prefill, a 5.4$\times$ reduction. Relative to 7B prefill, the corresponding reductions are 14.5$\times$ and 16.7$\times$. The advantage increases with context length, consistent with the goal of avoiding repeated prompt processing.

\begin{table}[t]
    \centering
    \small
    \setlength{\tabcolsep}{3pt}
    \caption{Within-family prefill and KV-handoff latency.}
    \label{tab:qwen-timing}
    \begin{tabular}{lrrr}
        \toprule
        \textbf{Context length} & \textbf{7B prefill} & \textbf{1.5B prefill} & \textbf{Handoff} \\
        \midrule
        8K--16K  & 500.6 ms & 158.7 ms & \textbf{34.5 ms} \\
        16K--32K & 898.7 ms & 288.3 ms & \textbf{53.8 ms} \\
        \bottomrule
    \end{tabular}
\end{table}

Table~\ref{tab:qwen-breakdown} further reports three instrumented components of the handoff path. Translation itself takes only 9.1--12.2~ms; peer copy and cache assembly add 5.7--8.1~ms. Together, these components account for 14.8~ms and 20.3~ms in the two context buckets, respectively. The remainder of the measured end-to-end handoff latency is not attributed by the current component-level instrumentation, indicating that runtime integration, synchronization, and other system overheads remain important optimization targets.

\begin{table}[t]
    \centering
    \small
    \setlength{\tabcolsep}{2.5pt}
    \caption{Instrumented components of within-family KV handoff.}
    \label{tab:qwen-breakdown}
    \begin{tabular}{lrrr}
        \toprule
        \textbf{Context length} & \textbf{Translation} & \textbf{Peer copy} & \textbf{Assembly} \\
        \midrule
        8K--16K  & 9.1 ms  & 3.3 ms & 2.4 ms \\
        16K--32K & 12.2 ms & 4.7 ms & 3.4 ms \\
        \bottomrule
    \end{tabular}
\end{table}

\subsection{Cross-Family Results}
\label{sec:cross-family-results}

\paragraph{Comparable-scale transfer: Qwen2.5-1.5B $\rightarrow$ Gemma-2-2B.}
We first isolate cross-family heterogeneity using models of comparable scale. Table~\ref{tab:qwen-gemma-efficiency} shows that handoff is consistently cheaper than native Gemma prefill, and its relative advantage grows with prompt length. Handoff reduces incremental latency by 45.25\% at 128 tokens, 59.90\% at 1K tokens, and \textbf{67.04\% at 4K tokens}. At 4K tokens, for example, the translated-cache path takes 59.897~ms, compared with 181.706~ms for native Gemma prefill.

\begin{table}[t]
    \centering
    \scriptsize
    \setlength{\tabcolsep}{2pt}
    \caption{Latency of Qwen2.5-1.5B $\rightarrow$ Gemma-2-2B KV sharing.}
    \label{tab:qwen-gemma-efficiency}
    \begin{tabular}{lrrrr}
        \toprule
        \textbf{Prompt} & \textbf{Qwen} & \textbf{Gemma} & \textbf{Handoff} & \textbf{Reduction} \\
        \textbf{length} & \textbf{prefill} & \textbf{prefill} & \textbf{latency} & \textbf{vs. Gemma} \\
        \midrule
        128 & 4.126 ms   & 5.492 ms   & 3.007 ms  & 45.25\% \\
        1K  & 30.684 ms  & 40.105 ms  & 16.081 ms & 59.90\% \\
        4K  & 136.972 ms & 181.706 ms & 59.897 ms & \textbf{67.04\%} \\
        \bottomrule
    \end{tabular}
\end{table}

The efficiency gains do not come at the cost of a uniformly degraded decoding state. Table~\ref{tab:qwen-gemma-ppl} shows that transferred-state perplexity is comparable to the native baselines across all horizons. At $H=1$, the handoff obtains the lowest perplexity (14.086), directly demonstrating that the translated cache provides a useful state at the first target-side decoding step. It also slightly outperforms native Gemma at $H=32$ and $H=128$. The pattern is not monotonic: handoff perplexity is higher at $H=16$ and $H=512$, with the largest gap from native Gemma occurring at $H=512$ (14.037 versus 13.574). These results show that the cross-family state is usable throughout decoding while revealing horizon-dependent interactions between the translated initial state and subsequent target-model computation.

\begin{table}[t]
    \centering
    \small
    \setlength{\tabcolsep}{3pt}
    \caption{Perplexity after Qwen2.5-1.5B $\rightarrow$ Gemma-2-2B KV transfer (lower is better).}
    \label{tab:qwen-gemma-ppl}
    \begin{tabular}{lccc}
        \toprule
        \textbf{Horizon} & \textbf{Qwen native} & \textbf{Gemma native} & \textbf{Handoff} \\
        \midrule
        $H=1$   & 14.721 & 14.384 & \textbf{14.086} \\
        $H=16$  & 14.126 & \textbf{13.842} & 14.517 \\
        $H=32$  & 13.998 & 13.731 & \textbf{13.642} \\
        $H=128$ & 13.884 & 13.612 & \textbf{13.584} \\
        $H=512$ & 13.847 & \textbf{13.574} & 14.037 \\
        \bottomrule
    \end{tabular}
\end{table}

\begin{table}[t]
    \centering
    \caption{Accuracy and latency for Llama3.1-70B $\rightarrow$ Qwen2.5-7B KV sharing.}
    \label{tab:llama-qwen-cross-family}
    \begin{tabular}{lcc}
        \toprule
        \textbf{Inference mode} & \textbf{Accuracy} & \textbf{Latency} \\
        \midrule
        Llama3.1-70B native & 44.0\% & 7,328 ms \\
        Qwen2.5-7B native & \textbf{45.7\%} & 899 ms \\
        Llama $\rightarrow$ Qwen KV handoff & 44.0\% & \textbf{138 ms} \\
        \bottomrule
    \end{tabular}
\end{table}
\paragraph{Large-to-small transfer: Llama3.1-70B $\rightarrow$ Qwen2.5-7B.}
We next introduce both a family change and a 10$\times$ parameter-scale difference. As reported in Table~\ref{tab:llama-qwen-cross-family}, the translated-cache path achieves 44.0\% accuracy, matching native Llama3.1-70B and trailing native Qwen2.5-7B by only 1.7 percentage points. Equivalently, it retains 96.3\% of the native target accuracy despite the architectural and scale mismatch.

The latency advantage is larger: handoff takes 138~ms, compared with 899~ms for native Qwen2.5-7B and 7,328~ms for native Llama3.1-70B. Reusing the existing Llama state is therefore 6.5$\times$ faster than target self-inference and 53.1$\times$ faster than source self-inference. Together with the comparable-scale Qwen-to-Gemma results, this experiment shows that KV transport remains effective under two distinct forms of cross-family heterogeneity: similar-scale models with different internal representations and a large-to-small pair that differs in both family and capacity.

\section{Conclusion}
This work investigates cross-model KV sharing as a mechanism for reusing previously computed contextual state across heterogeneous language models. By extending KV-cache reuse beyond the conventional same-model setting, the proposed approach enables a source model's contextual representation to be translated into a form that can support inference in a different target model. Our experimental results provide initial evidence that this form of state transfer is feasible both within a model family and across distinct model families, including settings with substantial differences in model scale, architecture, and internal representation. Across the evaluated configurations, translated KV states preserve meaningful downstream utility while reducing the need for redundant target-side prefill computation. These findings suggest that KV caches can be viewed not only as model-local execution artifacts, but also as transferable computational state. More broadly, the results motivate the notion of \emph{context mobility}, in which previously computed context can retain value as execution moves across models and inference stages. While further evaluation is needed to characterize generalization across additional architectures, workloads, context lengths, and deployment environments, the current results establish a promising foundation for cross-model KV reuse as a systems abstraction for efficient heterogeneous LLM and multi-agent inference.
\bibliographystyle{unsrt}
\bibliography{bib}

\end{document}